\documentclass[conference,final]{IEEEtran}
\IEEEoverridecommandlockouts

\newcommand{\Comment}[1]{}

\usepackage{cite}
\usepackage{amsmath,amssymb,amsfonts,amsthm}
\usepackage{algorithmic}
\usepackage{graphicx}
\usepackage{textcomp}
\usepackage{xcolor}
\usepackage{xspace}
\usepackage{enumitem}

\newtheorem{theorem}{Theorem}

\def\x{{\mathbf x}}

\def\R{\rm I\!R}

\def\BibTeX{{\rm B\kern-.05em{\sc i\kern-.025em b}\kern-.08em
    T\kern-.1667em\lower.7ex\hbox{E}\kern-.125emX}}

\providecommand{\abs}[1]{\lvert#1\rvert}

\definecolor{light}{gray}{.9}

\newcommand{\offset}{\textsc{Offset}\xspace}

\newcommand{\nxt}{\mathrm{next}}
\newcommand{\prv}{\mathrm{prev}}

\DeclareMathOperator*{\argmin}{argmin}

\DeclareMathOperator{\logit}{logit}

\DeclareMathOperator{\msqr}{msqr}

\usepackage{etoolbox} \makeatletter \patchcmd{\maketitle}{\@copyrightspace}{}{}{} \makeatother

\begin{document}

\title{Mitigating Divergence of Latent Factors via Dual Ascent for Low Latency Event Prediction Models}

\author{\IEEEauthorblockN{Alex Shtoff}
\IEEEauthorblockA{
\textit{Yahoo Research}\\
ashtoff@verizonmedia.com}
\and
\IEEEauthorblockN{Yair Koren}
\IEEEauthorblockA{
\textit{Yahoo Research}\\
yairkoren@verizonmedia.com}}


\maketitle

\begin{abstract}

Real-world content recommendation marketplaces exhibit certain behaviors and are imposed by constraints that are not always apparent in common static offline data sets. One example that is common in ad marketplaces is swift ad turnover. New ads are introduced and old ads disappear at high rates every day. Another example is ad discontinuity, where existing ads may appear and disappear from the market for non negligible amounts of time due to a variety of reasons (e.g., depletion of budget, pausing by the advertiser, flagging by the system, and more). These behaviors sometimes cause the model's loss surface to change dramatically over short periods of time. To address these behaviors, fresh models are highly important, and to achieve this (and for several other reasons) incremental training on small chunks of past events is often employed. These behaviors and algorithmic optimizations occasionally cause model parameters to grow uncontrollably large, or \emph{diverge}. In this work present a systematic method to prevent model parameters from diverging by imposing a carefully chosen set of constraints on the model's latent vectors. We then devise a method inspired by primal-dual optimization algorithms to fulfill these constraints in a manner which both aligns well with incremental model training, and does not require any major modifications to the underlying model training algorithm.

We analyze, demonstrate, and motivate our method on \offset, a collaborative filtering algorithm which drives Verizon Media (VZM) native advertising, which is one of VZM's largest and faster growing businesses, reaching a run-rate of many hundreds of millions USD per year. Finally, we conduct an online experiment which shows a substantial reduction in the number of diverging instances, and a significant improvement to both user experience and revenue.

\end{abstract}

\begin{figure*}[t]
\centering
\includegraphics[width=.6\textwidth]{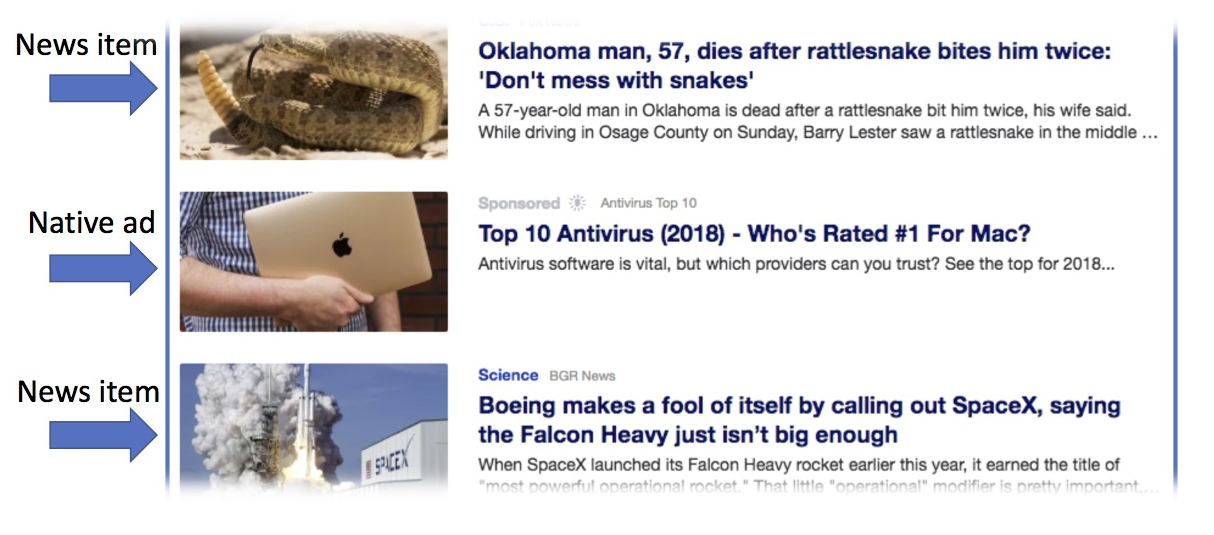}
\caption{Verizon Media native ads on Yahoo! homepage.}
\label{fig:gemini native}
\end{figure*}

\section{Introduction}\label{sec:Introduction}
The Verizon Media (VZM) native ad marketplace\footnote{See https://gemini.yahoo.com/advertiser/home} (previously known as \textit{Yahoo Gemini native}) serves users with native ads that are rendered to resemble the surrounding native content (see Figure \ref{fig:gemini native}). In contrast to the search-ads marketplace, users' intent during page (or site) visits are generally unknown.

Launched seven years ago and operating with a yearly run-rate of several hundreds of millions of USD, VZM native is one of VZM's most prominent and fastest growing businesses. With billions of ad views (impressions) daily, and an inventory of hundreds of thousands of active ads, this system performs real-time auctions that take into account user preferences, ad targeting, and budget considerations.

In order to rank the native ads for the incoming users and their specific context according to the cost per click (CPC) price type, the expected revenue of each ad is computed as a product of the advertiser's bid and the predicted click probability (pCTR). The pCTR is calculated using models that are continually updated by \offset\ - a feature enhanced collaborative-filtering (CF) based event prediction latent factor model \cite{aharon2013off}. 

In content marketplaces (e.g., articles, videos, ads, etc.), the set of users and content items may change abruptly and dramatically over time. Ad marketplaces are particularly susceptible since both the number of ads and their turnover in the marketplace at any given time is significant. Ad marketplaces may easily contain hundreds of thousands and sometimes millions of ads. New ads constantly flow in and old ads similarly flow out. Moreover, existing ads may be turned on or off in a heartbeat (e.g., due to budget being depleted or renewed). This discontinuous behavior is also propagated to the user side since some of the most prominent user features are the users' historical interactions with the various ads. To address these issues, models must swiftly adapt to the changing trends in the marketplace, and online incremental training is a natural choice. Moreover, space and regulation considerations often require deleting old data. If this old data is valuable for capturing long term user preference patterns, incremental training becomes a requirement. As an example, GDPR regulation requires deleting certain types of data that is older than 30 days whereas we've observed that our models require two to three months of training data in order to perform optimally.

To address these issues, we employ \offset in cycles, where in each cycle the algorithm trains on a chunk of training samples using multiple instances running in parallel, where each instance is configured with a different setting of hyper-parameters, e.g., step size. In particular, we train \offset using a variant of the Adagrad \cite{duchi2011adaptive} algorithm. At the end of each cycle, the best performing model is selected and saved for ad ranking. This best model is also used as an initial point for the next training cycle. The chunk training is performed in near real time fashion where the difference between the time of an event and its usage within a production model used to rank ads may be as low as several minutes. The mechanism is described in detail in \cite{aharon2017adaptive}. 

Consequently, we achieve two objectives: we adapt the model and its hyper-parameters to the changing trends, and more importantly, we adapt the ad ranking system to the changing trends by frequently feeding it with a fresh model. Indeed, it is our experience, as well as the experience of others (see \cite{debt, trends, zhang2020retrain}), that model freshness is of paramount importance to the performance of content recommendation systems deployed in a changing environment. Moreover, it is our experience that the parameters of the models as well as their optimal hyper-parameters change over time, sometimes significantly over short periods. In such settings it is sub optimal to deploy a model trained on older data or with a-priori chosen hyper-parameter settings.

The optimization algorithms which are often employed when training models are not guaranteed to converge, or even stay in a bounded region, for all hyper-parameter settings. Indeed, we observed that for some hyper-parameter settings, the latent factors of our model grow uncontrollably large, causing the corresponding training instance to fail to produce a meaningful model and subsequently be discarded by our hyper-parameter tuning algorithm. When a large enough percentage of the instances are discarded, the pool of models to choose the best model from is substantially reduced, and consequently the performance of the chosen best model for ad ranking is impaired. 

Thus, in this work we aim to improve the robustness of the \emph{training algorithm} to the choice of hyper-parameters. We show, by empirically analyzing the latent factors of the model, that only a few latent factor models are affected by bad hyper-parameter choice, and that it is indeed possible to prevent most instances from diverging without impairing the model's performance. We then show how to achieve the goal by imposing a carefully chosen bound on the model's latent vector norms using a mechanism that aligns well with incremental model training. We employ the duality theory in optimization to decouple model training from the bound constraint enforcement, and devise a method based on the well-known idea of primal-dual optimization algorithms such that the only change to the underlying model training algorithm is a different regularizer. We specialize the discussion to \offset, but the ideas we present here will likely be useful for any factorization based model. To the best of our knowledge, the study of divergent behavior when training collaborative filtering models is new, and our primal-dual approach for divergence mitigation is novel.

The rest of the paper is organized as follows. After reviewing related work in Section \ref{sec:related_work}, in Section \ref{sec:background} we describe details about the \offset model, its training algorithm, and how divergence is defined, detected and handled during training. Then, we take a deeper look into the behavior of a diverging model in Section \ref{sec:study} from both theoretical and empirical perspectives, and draw conclusions that lead to our method, described in Section \ref{sec:mitigation}. Finally, we demonstrate our method's improved performance in Section \ref{sec:evaluation}.
\section{Related work}\label{sec:related_work}
The divergent behavior we study in this work boils down to failure to keep the iterates of the optimization algorithm that trains the model in a bounded region of space. To the best of our knowledge, studying and mitigating divergent behavior has received little attention in the literature, and thus the body of research devoted the subject is quite modest.

It is well known (e.g. Example 1 in \cite{asi_duchi_2018}), that when the trained loss functions grow faster than a quadratic polynomial, divergent behavior under certain hyper-parameter choices is inevitable, even for \emph{convex} functions. Several works, such as \cite{boyd_spi, asi_duchi_2020, davis_drusvyatskiy}, propose algorithms to mitigate the issue for the stochastic optimization setting. When certain assumptions are met, the proposed algorithms ensure that the parameters of the model indeed stay in a bounded region for any step-size choice. These algorithms are mainly useful when little resources for fine hyper-parameter tuning are available, however, their usefulness is limited in a scenario where significant model optimization is required. Indeed, the numerical experiments in \cite{asi_duchi_2018} show that models trained with classical algorithms using tuned hyper-parameters often perform better. Moreover, our scenario of interest does not fit into the stochastic setting where the training data is sampled from a stationary distribution. The contrary is true - we operate in a non continuous, constantly changing environment, and thus we train our models in an \emph{online} fashion.  Our proposed primal-dual approach aligns well with online training, \emph{without} introducing significant changes the underlying training algorithm. The above is an important property when using an algorithm which has passed the test of time, and proved to be very reliable for the task at hand.

A different but related line of work is the convexification of latent factor models, such as \cite{cvx_compact_fm, cvx_fm_book, cvx_fm_toxico}. In these bodies of work, the decision variables of the resulting optimization problem are inherently bounded by construction, since the convexification procedure itself directly represents a matrix of inner products of the latent factors, instead of the factors themselves, and maintains a low rank of this matrix by imposing a threshold on its nuclear norm\footnote{For a matrix $A$, its nuclear norm $\|A\|_*$ is the sum of its singular values.}. These methods, despite promising to achieve a globally optimal solution in terms of the training loss, are computationally expensive and do not fit the low latency scenario, where the amount of data is vast, and speed is of the essence, due to the importance of the freshness of the deployed model. 

\section{Background}\label{sec:background}
\subsection{Verizon Media Native}\label{sec: Gemini Native}
The VZM native ads platform serves billions of ad impressions to several hundreds of millions of users world wide, using a native ad inventory of hundreds of thousands of active ads. Native ads resemble the surrounding page items, are considered less intrusive to users, and provide a better user experience in general (see Figure \ref{fig:gemini native}).
The online serving system is comprised of a massive Vespa deployment \cite{vespa}, VZM's open source elastic search solution, augmented by ads, budget and model training pipelines. The Vespa index is updated continually with ad and budget changes, and periodically (e.g., every $15$ minutes) with model updates resulting from each training cycle. The VZM native marketplace serves several ad price-types including CPC (cost-per-click), oCPC (optimizing for conversions), CPM (cost-per-thousand impressions), CPV (cost-per-video-view), and also includes RTB (real-time biding) in its auctions. 

\subsection{The \offset Click-Prediction Algorithm}\label{sec:offset}
The algorithm driving Verizon media native models is \offset (One-pass Factorization of Feature Sets): a feature enhanced collaborative-filtering (CF)-based ad click-prediction algorithm \cite{aharon2013off}, which resembles a factorization machine \cite{rendle_fm}. The \textit{predicted click-through-rate} (pCTR) of a given user $u$ and an ad $a$ according to \offset is given by
\begin{equation}\label{eq: pCTR}
    \mathrm{pCTR}(u,a) = \sigma(\logit(u, a))\in [0,1]\ ,
\end{equation}
where $\sigma(x)=\left(1+e^{-x}\right)^{-1}$ is the \emph{Sigmoid function}, and 
\begin{equation}\label{eq: score}
	\logit(u, a) = b+\nu_{u}^T\nu_{a}\ ,
\end{equation}
where $\nu_{u},\nu_{a}\in \R^N$ denote the user and ad latent factor vectors respectively, and $b\in \R$ denotes the model bias. The product $\nu_{u}^T\nu_{a}$ reflects the affinity score of user $u$ towards ad $a$, where a higher score translates into a higher pCTR. Both ad and user vectors are constructed using their features' vectors, which enable dealing with data sparsity and cold start issues (ad CTR is quite low in general). For ads, we use a simple summation between the vectors of their features (e.g., unique creative id, campaign id, advertiser id, ad categories, etc.), all in dimension $N$. The combination between the different user feature vectors is more complex in order to allow non-linear dependencies between feature pairs. 
\begin{figure}[!t]
\centering
\includegraphics[width=.9\columnwidth]{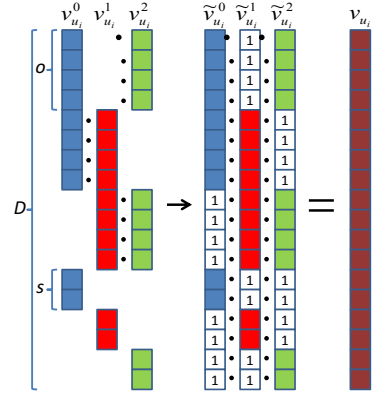}
\caption{Example of a user latent factor vector construction for $o=4,\ s=2$ and $K=3$ features (i.e., age, gender, and geo). Assume user $u_i$ is a 30 year old female from NY city (i.e., $\nu_{u_i}^0$ blue - age 30 , $\nu_{u_i}^1$ red - female, and $\nu_{u_i}^2$ green - NY city). After filling the empty entries with 1s, the three vectors (i.e., $\widetilde{\nu}_{u_i}^0$, $\widetilde{\nu}_{u_i}^1$, $\widetilde{\nu}_{u_i}^2$) are multiplied entry wise to get the final user $u_i$ vector $\nu_{u_i}$ (in brown).}
\label{fig: user vector construction}
\end{figure}
The user vectors are composed of their $K$-feature latent vectors $v_k\in \R^n$ (e.g., age, gender, geo, etc.). In particular, $o$ entries are allocated to each pair of user feature types, and $s$ entries are allocated to each feature type vector alone.
The dimension of a single feature value vector is therefore $d=(K-1)\cdot o + s$, whereas the dimension of the combined user vector is $N={K\choose 2} \cdot o + K \cdot s$. An illustration of this construction is given in Figure~\ref{fig: user vector construction}. The model's parameters are the individual vectors associated with each feature value, and the model's bias $b$. 

As with factorization machines, the advantage over the standard CF approach is that the model includes only $O(K)$ feature latent vectors (one for each feature value, e.g., 3 for gender - male, female and unknown) instead of hundreds of millions of unique user latent vectors.

To learn the model parameters $\theta$, \offset minimizes the binary cross-entropy loss (or log-loss) of the training data set $\mathcal{T}$ (i.e., past impressions and clicks) using one-pass variant of Adagrad \cite{duchi2011adaptive}. The loss function is as follows:
\begin{equation*}
\argmin_{\theta} \quad \sum_{(u,a,y)\in \mathcal{T}} \mathcal{L}(u,a,y)\ ,
\end{equation*}
where
\begin{multline}
\label{eq: logloss}
\mathcal{L}(u,a,y)=-(1-y)\log\left(1-\mathrm{pCTR}(u,a)\right) \\ -y \log \mathrm{pCTR}(u,a)+\frac{\lambda}{2} \|\theta_{u,a}\|_2^2 ,
\end{multline}
$y \in \{0,1\}$ is the click indicator (or binary label) for the event involving user $u$ and ad $a$, $\theta_{u,a}$ are the latent vectors corresponding to the feature values of user $u$ and ad $a$, and $\lambda$ is a global $\ell_2$ regularization parameter. 
For each training event $(u,a,y)$, \offset updates its relevant model parameters by the gradient-based step
\[
\theta \gets \theta-\eta \odot \nabla_\theta\mathcal{L}(u,a,y)\ ,
\]
where $\nabla_\theta\mathcal{L}(u,a,y)$ is the gradient of the loss function w.r.t $\theta$, and $\odot$ is a component-wise product between two vectors. The parameter dependent step size is given by
\[
\eta = \frac{\eta_0}{\alpha+\left(\sum_{(u,a,y) \in S}\abs{\nabla_\theta \mathcal{L}(u,a,y)}\right)^\beta}\ ,
\]
where $\eta_0$ is the initial step-size, $\alpha, \beta\in\R_+$ are the parameters of our variant of the adaptive gradient (AdaGrad) algorithm \cite{duchi2011adaptive}, and $S$ is the set of training samples observed so far. The \offset algorithm uses an online approach where it continually updates its model parameters with each chunk of new training events (e.g., every 15 minutes for the click model).

The \offset algorithm includes an adaptive online hyper-parameter tuning mechanism \cite{aharon2017adaptive}. This mechanism takes advantage of our parallel hardware architecture to run many instances of \offset in parallel in each training cycle, each instance with its own set of hyper-parameters, and strives to tune \offset hyper-parameters, such as $\eta_0, \alpha$, and $\beta$ above, to match the varying marketplace conditions. An instance is defined to be diverging, and consequently is aborted and discarded from consideration by our hyper-parameter tuning algorithm, if for some latent vector $v$ we have
\[
\|v\|_\infty \geq \tau,
\]
where $\tau$ is some predefined threshold set based on previous experience, and $\|v\|_\infty$ is the maximum absolute value of any component of $v$. In our system we use $\tau = 15$, since form our experience, in that case the instance will almost certainly continue grownig uncontrollably large and fail to produce a meaningful model.

Finally, we point out that the set of latent vectors in the model changes over time as well, since new feature values appear in the stream of training events, while old feature values disappear from the stream. For example, when a new ad appears which has not been previously encountered, a corresponding latent vector is created and initialized with normally-distributed elements with a zero mean and a small standard deviation. When an ad is no longer encountered by our training algorithm for a predefined period of time, its latent vector is removed from the model. Thus, the dimension of the model parameters vector $\theta$, which can be seen as a concatenation of all of the latent vectors of the model, changes over time as well. As mentioned in a previous section, these changes can be significant in quantity, abrupt, and induce a non continuous dynamic to the model.

Other components of \offset, such as its weighted multi-value feature \cite{arian2019feature}, and similarity weights used for applying ``soft'' recency and frequency rules\footnote{How frequently and how recently a user was presented with a certain ad.} \cite{aharon2019soft}, are not presented here for the sake of brevity.

\section{Analytical and empirical study}\label{sec:study}
An integral part of solving the divergence issue is gaining a deeper insight from both empirical and analytical perspectives. We first take the analytical perspective by \emph{qualitatively} inspecting the updates of any training algorithm based on gradient updates with coordinate-wise step sizes, such as AdaGrad \cite{duchi2011adaptive}.
Denoting the composition of the Sigmoid function with the binary cross-entropy loss corresponding for the label $y$ by $\Phi_y$, the loss in \eqref{eq: logloss} can be equivalently written as
\begin{equation}
    \label{eq:loss_composition}
    \Phi_y(\logit(u, a)) + \frac{\lambda}{2} \|\theta_{u,a}\|_2^2,
\end{equation}
Consider a training sample $(u, a, y)$ and look at the gradient of the loss \eqref{eq:loss_composition} with respect to some vector $z$ corresponding to a feature of the ad $a$:
\[
\nabla_z \Phi_y(\logit(u, a)) = \Phi_y'(\logit(u, a)) \cdot \nu_u.
\]
When training the model, the algorithm will update $z$ according to the rule:
\[
z^{\mathrm{next}} = z^{\mathrm{curr}} - \eta \odot \Phi_y'(\logit(u, a)) \cdot \nu_u,
\]
where $\eta$ is some vector of coordinate-wise step sizes which depends on the training algorithm of choice. From the above we conclude that if some component of $\nu_u$ is large enough and has the appropriate sign, the gradient update will substantially increase the magnitude of the corresponding component of $z$. Symmetrically, considering a vector $w$ corresponding to some feature of the user $u$ produces a similar argument - if some component of the ad vector $\nu_a$ or an overlapping user feature vector turns out to be large enough and of the appropriate sign, the corresponding component of $w^{\mathrm{next}}$ will substantially increase in magnitude. Consequently, divergence is \emph{contagious}: it's enough for a few vector components to start growing beyond control to 'contaminate' the entire model.

\begin{figure}
    \centering
    \includegraphics[width=\columnwidth]{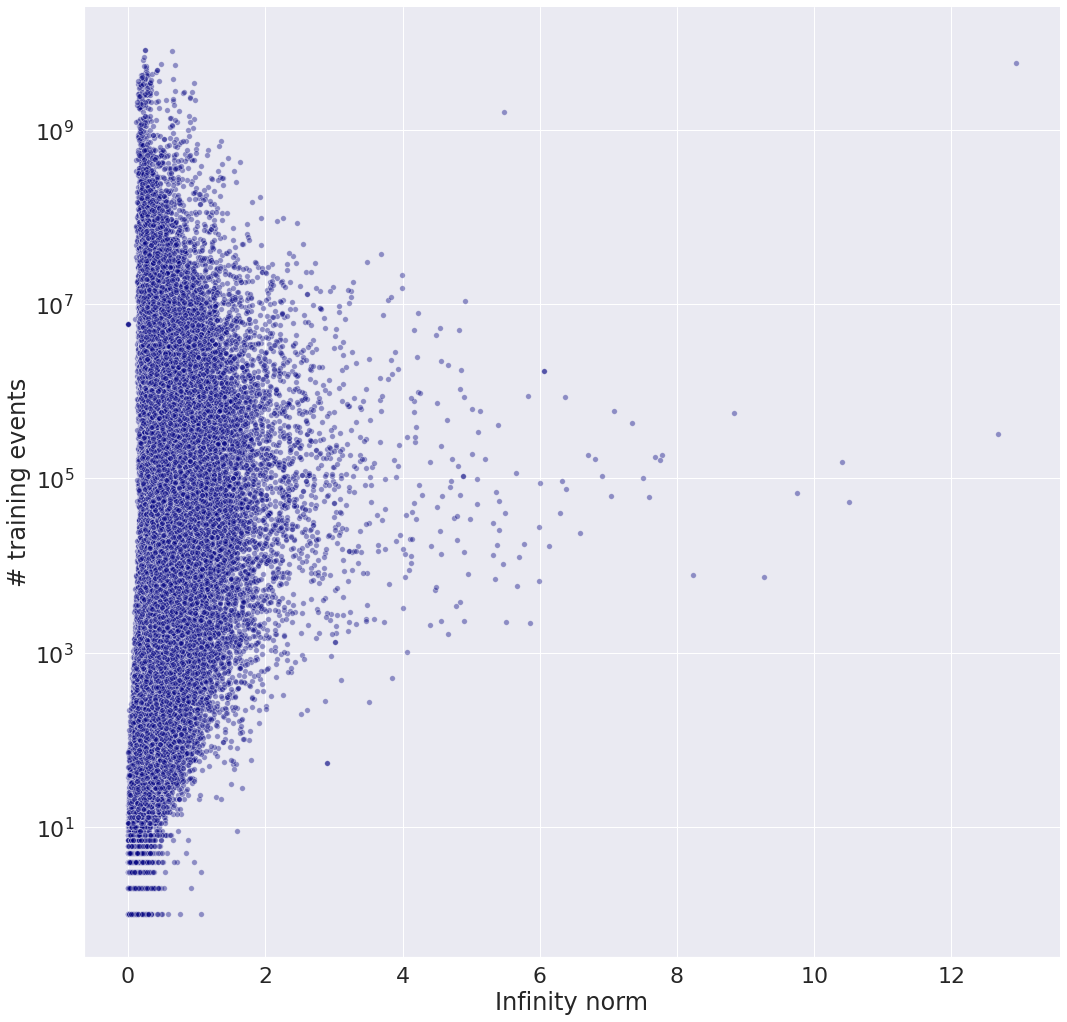}
    \caption{Vector infinity norms of latent vectors before training abort. Each vector is plotted as a point whose $x$ coordinate is its infinity norm, while its $y$ coordinate is the number of training samples which caused that vector to be updated.}
    \label{fig:vector_norms}
\end{figure}

To study the phenomenon empirically, we located diverging instances and plotted $\|\nu\|_\infty$ for all the latent vectors $\nu$ versus the number of training samples which caused that vector to be updated, just before the training in that cycle was aborted. A typical plot can be seen in Figure \ref{fig:vector_norms}, where each point is a vector whose $x$ coordinate is its infinity norm, while its $y$ coordinate is the number of training samples which caused that vector to be updated. Its apparent that new vectors which were not updated by many training samples, or were not updated frequently, have small norms. That's not a surprise - newly added vectors are initialized to have small random elements. Among the more mature vectors, we see only a few vectors with exceptionally large norms, but still most vectors retain a small norm.

From these two observations we draw a simple conclusion: (a) since norm growth is contagious, it's essential to maintain vector norms below a certain threshold to mitigate divergence. (b) Imposing such a threshold without hurting the model's performance seems to be possible, since only a small fraction of the model's vectors will be affected. 

\section{Mitigation through a constrained optimization approach}\label{sec:mitigation}
To make sure our vectors have small elements, we impose a constraint on their Euclidean norms, or equivalently, their mean-squared element:
\[
\msqr(x) = \frac{1}{d} \|x\|_2^2, \qquad \text{where } x \in \mathbb{R}^d.
\]
Thus, instead of training by minimizing \eqref{eq:loss_composition}, we train by striving to solve the optimization problem
\begin{equation}
\label{eq:constrained_form}
\begin{aligned}
\min_{\theta} &\quad \sum_{(u, a, y) \in S} \Phi_y(\logit(u, a)),  \\
\text{subject to} &\quad \msqr(\theta_v) \leq \rho, &&\quad \forall v \in \mathcal{V}_{u,a}
\end{aligned}
\end{equation}
where $\theta_v$ is the latent vector corresponding to the feature value $v$, and $ \mathcal{V}_{u,a}$ is the set of latent vectors corresponding to the features of user $u$ and ad $a$. 

Recall that divergence is detected and analyzed empirically in terms of $L_\infty$ norms of the latent vectors, while our constraints impose a limit on their squared $L_2$ norm. There are two primary reasons. First, from a practical perspective, it's easier to work with the squared Euclidean norm due to its differentiability. Second, norms in a finite dimensional real space are equivalent up to a constant, and in particular we have
\[
\frac{1}{\sqrt{d}}\|x\|_2 \leq \|x\|_\infty \leq \|x\|_2, \qquad \text{where } x \in \mathbb{R}^d.
\]
Therefore, making the $L_2$ norm smaller potentially also induces a smaller $L_\infty$ norm, and our numerical experiments in Section \ref{sec:evaluation} show that it is indeed the case. And finally, we formulate using the mean squared element instead of equivalently using the squared Euclidean norm, since there is a good heuristic for discovering the upper bound $\rho$ for that case, which appears later in this section.

Observe also that the regularization term has been removed from the loss, since imposing constraints is, in itself, a form of regularization. As a by-product, we also make the hyper-parameter tuning mechanism more efficient, since we eliminate the global regularization coefficient $\lambda$ in \eqref{eq:loss_composition} from the set of hyper-parameters to be tuned.

In theory, solving \eqref{eq:constrained_form} can be performed by replacing any optimization algorithm with its projected variant, e.g. gradient projection \cite{gradient_projection_goldstein}: after each gradient step, vectors whose mean squared element is above the threshold are divided by a constant to normalize their mean squared element. However, projected algorithms are prohibitive in an incremental training setting where models are not trained from scratch, but instead continuously initialized from an existing model and trained over a new chunk of data. The projection operator will substantially modify the latent vectors having large norms, and it from our experience it takes a prohibitively long time for a new model to converge when initialized from a previous model which does \emph{not} impose constraints. Consequently, we impose constraints using a considerably less aggressive mechanism, inspired by primal-dual optimization algorithms. The remainder of this section is devoted to describing the algorithm in detail, and to a heuristic for discovering a good upper bound $\rho$ for the formulation \eqref{eq:constrained_form}.

\subsection{A na\"\i ve approach}\label{sec:naive}
Before diving deeper, we first introduce a simple approach that demonstrates the idea, although we found it to be non-effective. It's described for two reasons. First, from a pedagogical perspective, it allows readers unfamiliar with duality to grasp the intuition behind our algorithm. Second, it demonstrates that a more systematic approach may be required, and provides some intuition about where improvements may be made.

We attempt to solve \eqref{eq:constrained_form} by replacing the the global regularization coefficient in the loss \eqref{eq:loss_composition} with a separate regularization coefficient for each latent vector. Formally, upon receiving the training sample $(u, a, y)$ we perform the following steps:
\begin{description}
    \item[Train] Perform a training step using the loss
    \[
        \Phi_y(\logit(u, a)) + \sum_v \mu_v \msqr(\theta_v)
    \]
    \item[Control] If $\msqr(\theta_v) > \rho$, update $\mu_v$ by multiplying it with some constant $\beta > 1$. Otherwise, divide $\mu_v$ by $\beta$.
\end{description}
If the parameter $\beta$, which controls the aggressiveness of the control mechanism, is small enough, the control mechanism will gradually increase the regularization coefficients of the vectors having a mean squared element above the threshold, thus causing the following training steps to reduce their norms. Alternatively, vectors whose mean squared element is below the threshold will have their regularization coefficient driven towards zero.

Unfortunately, this simple control mechanism did not work in practice. We were not able to find $\beta$ which dramatically reduces the number of retained instances, without worsening the resulting model's performance. Our conjecture was that the increase or decrease of the coefficients $\mu_v$ should be somehow related to the severity of the constraint violation, i.e. a larger $\msqr(\theta_v) - \rho$ should result in a more aggressive update, and duality provides a systematic way of deriving such update formulas.

\subsection{A short review of duality}
Duality is a central component of modern optimization theory and practice, and is described in most standard optimization textbooks, such as \cite{beck_nlopt}. To make this paper self-contained and accessible, we give a short review, and authors who are familiar should skip to the following subsection (Section \ref{sec:duality_approach}).

Suppose we are given an arbitrary constrained optimization problem
\begin{equation}
\label{eq:primal}
\begin{aligned}
\min_x &\quad f_0(x), \\
\text{subject to} &\quad f_j(x) \leq 0 && j = 1, \dots, m.
\end{aligned}
\end{equation}
We define the following unconstrained counterpart defined for $\mu \in \mathbb{R}^d$ with $\mu_j \geq 0$:
\[
q(\mu) = \min_x \left\{ L(x, \mu) \equiv f_0(x) + \sum_{j=1}^m \mu_j f_j(x) \right \}.
\]
Namely, we assign a 'price' $\mu_j \geq 0$ paid for the violation of the constraint $f_j(x) \leq 0$. For each vector of prices $\mu$ there is an optimal value $q(\mu)$. The function $L$, which serves as the minimization objective of the unconstrained problem is called the \emph{Lagrangian} associated with problem \eqref{eq:primal}.

A well known result is that $q(\mu)$ provides a lower-bound on the optimal value of the problem \eqref{eq:primal}. The \emph{dual problem} associated with \eqref{eq:primal} is concerned with finding the "best" lower bound, namely:
\[
\max_\mu \quad q(\mu) \quad \text{subject to} \quad \mu \geq 0.
\]
The original problem \eqref{eq:primal} is called the \emph{primal} problem. A well known result in convex analysis implies that under convexity the optimal values of both problems coincide. Formally,
\begin{theorem}[Strong duality]
Suppose that the functions $f_0, \dots, f_m$ in \eqref{eq:primal} are convex, the optimal value of \eqref{eq:primal} is finite, and there exists $\hat{x}$ such that $f_j(\hat{x}) < 0$ for all $j = 1, \dots, m$. Then,
\[
\max_\mu \{ q(\mu) : \mu \geq 0 \} = \min_x \{ f_0(x) :f_j(x) \leq 0, \, j = 1, \dots, m \}
\]
\end{theorem}
A proof, and extensive material on convex duality in particular and optimization theory in general can be found in, for example, \cite{beck_nlopt}.

As a consequence of strong duality, the idea of maximizing the dual and minimizing the primal gave rise to a vast variety of algorithms for convex optimization, called primal-dual methods, which eventually boil down to the idea of iteratively performing a pair of steps:
\begin{description}
    \item[Primal descent] Update $x$ by performing a \emph{descent} step on a variant of the Lagrangian $L(x, \mu)$, assuming $\mu$ is constant.
    \item[Dual ascent] Update $\mu$ by performing an \emph{ascent} step on a variant of the Lagrangian $L(x, \mu)$, assuming $x$ is constant.
\end{description}
Prominent examples include the classical dual ascent and augmented Lagrangian  methods \cite{dual_ascent, alm1, alm2}, the primal-dual methods by Nesterov \cite{nesterov2009primal} and by Zhu and Chan \cite{zhu_chan}. The latter  was improved by Chambolle and Pock in \cite{chambolle_pock}, and generalized in \cite{chambolle2016ergodic}.

Our problem of interest \eqref{eq:constrained_form} is non-convex, and even if it were, we are not really minimizing the loss over a given training set, but rather training on an infinite stream of samples. Thus, we cannot employ these algorithms as reliable building blocks with proven convergence guarantees. However, they do provide valuable insight into the design of powerful methods which work well in practice. 

\subsection{A duality-based approach}\label{sec:duality_approach}
Our method is a variant of the classical dual ascent method \cite{dual_ascent}. The Lagrangian corresponding to our constrained formulation \eqref{eq:constrained_form} is
\[
L(\theta, \mu) = \sum_{(u, a, y) \in S} \Phi_y(\logit(u, a)) + \sum_v \mu_v \left( \msqr(\theta_v) - \rho \right).
\]
Recalling that $\msqr$ is the mean-squared element function. Our heuristic performs a primal descent and a dual ascent step for each training sample.
\subsubsection{Primal descent}
We employ our regular training algorithm on the given training sample, where the Lagrangian is the training loss. Since primal descent minimizes over $\theta$ and treats $\mu$ as a constant, it is equivalent to training with the loss
\begin{multline*}
\sum_{(u, a, y) \in S} \Phi_y(\logit(u, a)) + \sum_v \mu_v \msqr(\theta_v) \\ =  \sum_{(u, a, y) \in S} \Phi_y(\logit(u, a)) + \sum_v \frac{\mu_v}{d_v} \|\theta_v\|_2^2,
\end{multline*}
where $d_v$ is the dimension of the latent vector $\theta_v$. Consequently, primal ascent boils down to training the model using a modified regularization term, which assigns a different coefficient to each latent vector.

\subsubsection{Dual ascent}
The partial derivatives of the Lagrangian with respect to the components of $\mu$ are
\[
\frac{\partial L}{\partial \mu_v} = \msqr(\theta_v) - \rho.
\]
An intuitive understanding of how dual ascent works can be obtained by analazying the simplest dual ascent algorithm - projected gradient ascent:
\begin{equation}\label{eq:proj_grad}
\mu_v^{\mathrm{next}} = \max\left(0, \mu_v^{\mathrm{prev}} + \beta (\msqr(\theta_v) - \rho)\right), 
\end{equation}
where $\beta > 0$ is the step size for the ascent step. Here, we perform a gradient ascent step, that is followed by a projection onto the non-negative numbers. If $\msqr(\theta_v) > \rho$, namely, we're violating our upper bound constraint, then the above update rule will increase $\mu_v$, and thus the next primal descent step will regularize it more aggressively. If $\msqr(\theta_v) \leq \rho$, then the above update rule will decrease $\mu_v$ towards zero, and thus the next primal descent step will regularize it less aggressively. Consequently, dual ascent is a kind of a \emph{control mechanism}, and its step-size $\beta$ affects its aggressiveness.

A major advantage of framing the mechanism to control latent vector norms as a dual optimization problem is the fact that it opens up the entire arsenal of optimization algorithms. Consequently, duality is a systematic framework for deriving update rules for $\mu$: every optimization algorithm implies a different control mechanism.

\subsubsection{Our ascent algorithm}
A well known result, e.g. \cite{beck2003mirror}, is that the projected gradient step \eqref{eq:proj_grad} can be alternatively written using the \emph{proximal formulation}
\begin{multline*}
\mu^{\nxt} = \argmin_{\mu} \Biggl\{  -\beta \underbrace{\langle \nabla_\mu L(x, \mu^{\prv}), \mu - \mu^{\prv} \rangle}_{\text{Alignment term}} \\ + \underbrace{\frac{1}{2} \|\mu - \mu^{\prv}\|_2^2}_{\text{Proximity term}} : \mu \geq 0 \, \Biggr\},
\end{multline*}
meaning "find $\mu \geq 0$ such that the update direction $\mu - \mu^{\prv}$ aligns with the gradient, while $\mu$ stays in close proximity to the previous iterate". The step-size $\beta$ balances between the two opposing forces. To achieve better convergence properties by adapting the algorithm to either the objective functions or the constraints, a common practice in optimization, e.g. see \cite{fukushima_bpg,beck2003mirror,chambolle2016ergodic}, is to use a more general measure of proximity instead of the squared Euclidean distance, and update $\mu$ using:
\begin{multline}
\label{eq:mirror_grad}
\mu^{\nxt} = \argmin_\mu \Biggl\{ \,
 -\beta \langle \nabla_\mu L(x, \mu^{\prv}), \mu - \mu^{\prv} \rangle \\ + D(\mu, \mu^{\prv}) : \mu \geq 0 \,
\Biggr\},
\end{multline}
where $D$ is a distance-like function called the Bregman divergence. Algorithms of this form are known as mirror ascent, as in \cite{beck2003mirror}, or Bregman gradient method, as in \cite{fukushima_bpg}. A rigorous presentation, including the definition of a Bregman divergence, can be found in \cite{beck2003mirror} and references therein, and an extension to the stochastic optimization setting on unbounded domains can be found in \cite{hanzely2018fastest}.

The squared Euclidean distance is one example of a Bregman divergence. In this work, our divergence of choice is
\[
D(u, v) = \sum_{i=1}^d \left[ u_i \ln(\tfrac{u_i}{v_i}) + v_i - u_i \right]
\]
with the convention that $0 \ln(0) \equiv 0$, for which the generic formula \eqref{eq:mirror_grad} reduces to the multiplicative update rule
\begin{equation}
\label{eq:entropic_ascent}
\mu_v^{\nxt} = \mu_v^{\prv} e^{\beta (\msqr(\theta_v) - \rho)}.
\end{equation}
Note, that this kernel results in a very simple update rule which does not involve any projection steps, and the resulting update formula is known by many names, such as the \emph{entropic gradient step} \cite{beck2003mirror}. As we shall see in the evaluation section, this choice works well in practice.

Intuitively, when $\msqr(\theta_v) > \rho$, the algorithm makes increases $\mu_v$ by multiplying it by a factor greater than one. Conversely, when $\msqr(\theta_v) < \rho$, the algorithm decreases $\mu_v$ towards zero by multiplying it by a factor smaller then one. Surprisingly, this approach, which works well in practice, bears similarity to the na\"\i ve approach described in Section \ref{sec:naive}.

\subsubsection{Summary}
To summarize, we modified the training algorithm such that for each training sample $(u, a, y)$ it performs the following steps:
\begin{description}
    \item[Primal descent] Update model parameters $\theta$ by train on the sample $(u, a, y)$ with the loss 
    \[
    \Phi_y(\logit(u, a)) + \sum_v \frac{\mu_v}{d_v} \|\theta_v\|_2^2.
    \]
    \item[Dual ascent] Update the dual variables $\alpha_v$ using the update formula in \eqref{eq:entropic_ascent}.
\end{description}

We would like to stress, again, that the approach above is not an algorithm with provable convergence guarantees, but rather a method inspired by primal-dual algorithms that turned out to be very effective in practice. Moreover, note that we did not specify \emph{how} the primal descent is employed - we can just keep the same training algorithm whose performance and reliability has been proved for our task at hand, e.g. \offset training stays unmodified in our case. Consequently, the approach is non-intrusive and generic enough for a variety of tasks which are solved by factorization models with their corresponding training algorithm.

\subsection{A heuristic for deriving bound $\rho$}
Our heuristic is based on an analysis of the sigmoid function $\sigma(t)=(1+\exp(-t))^{-1}$, and making some assumptions about the distribution of mass between the user and the ad side. In practice, our heuristic turned out to produce satisfactory results, and may serve as the basis for similar heuristics adapted to other factorization machines. We present it here, since a similar line of thought may be useful for deriving similar upper bounds for other factorization-machine based models.

The click probabilities are computed by composing the sigmoid function $\sigma(t)$ onto $\logit(u, a)$. Observe that if $t \notin [-12, 12]$ then $\sigma(t) \notin [10^{-5}, 1-10^{-5}]$. Therefore, if $\logit(u, a)$ lies in the interval $[-12, 12]$, we cover the entire range of plausible click probabilities. Consequently, our heuristic begins from requiring that
\begin{equation}
\label{eq:vec_product_req}
    |\langle \nu_u, \nu_a \rangle| \leq 12.
\end{equation}
Using the Cauchy-Schwartz inequality, namely, 
\[
    |\langle x, y \rangle| \leq \|x\|_2 \|y\|_2,
\]
we conclude that inequality \eqref{eq:vec_product_req} is ensured when
\begin{equation}
\label{eq:norm_prod_req}
    \| \nu_u \|_2^2 \|\nu_a\|_2^2 \leq 12^2.
\end{equation}
Let $t>0$ be some parameter describing the "division of mass" between the user and the ad side, and it will be determined later. Using this parameter, inequality \eqref{eq:norm_prod_req} is ensured if we require that
\[
    \|\nu_u\|_2 \leq \frac{12}{1+t}, \qquad 
    \|\nu_a\|_2 \leq \frac{12 t}{1+t}.
\]
The above can be equivalently written as
\begin{equation}
    \label{eq:sep_norm_req}
    \|\nu_u\|_2^2 \leq \left(\frac{12}{1+t}\right)^2, \qquad 
    \|\nu_a\|_2^2 \leq \left(\frac{12 t}{1+t}\right)^2,
\end{equation}
Denoting the length of the user and ad vectors by $N$, equation \eqref{eq:sep_norm_req} can be written as:
\begin{equation}
    \label{eq:sep_msqr_req}
    \msqr(\nu_u) \leq \frac{1}{N} \left(\frac{12}{1+t}\right)^2, \quad
    \msqr(\nu_a) \leq \frac{1}{N} \left(\frac{12 t}{1+t}\right)^2
\end{equation}
Since $\nu_u$ is constructed from products of user feature vectors in the overlapping part, we make a rough approximation by discarding the independent part of the user vectors and assuming that the components of the overlapping vectors each contribute roughly a square root of their product, and thus require that for each user feature vector $z$ we have
\begin{equation}
\label{eq:user_msqr_req}
\msqr(z) \leq \sqrt{\frac{1}{N} \left(\frac{12}{1+t}\right)^2} = \frac{1}{\sqrt{N}} \frac{12}{1+t}.
\end{equation}

\begin{figure}
    \centering
    \includegraphics[width=.8\columnwidth]{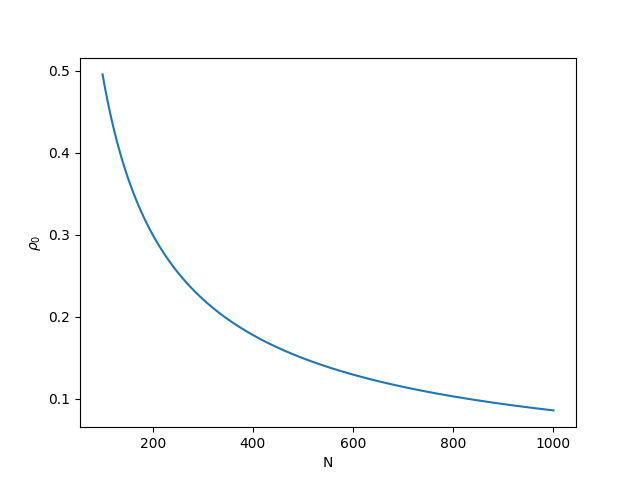}
    \caption{The value of the heuristic mean-squared element upper bound, as a function of the model length.}
    \label{fig:rho_zero}
\end{figure}

To summarize, equation \eqref{eq:sep_msqr_req} provides a bound on the mean squared element of an ad vector, whereas equation \eqref{eq:user_msqr_req} provides a similar bound for user feature vectors. We would like both bounds to be to the same value $\rho$, and therefore we solve the equation
\[
\frac{1}{\sqrt{N}} \frac{12}{1+t} = \frac{1}{N} \left(\frac{12 t}{1+t}\right)^2,
\]
which is equivalent to a quadratic equation. Having computed $t$, we substitute it into for example, its left hand side, and obtain the \emph{heuristic} mean-squared element bound $\rho_0$:
\[
\rho_0 = \frac{288}{24 \sqrt{N} + N^{\frac{3}{4}} \sqrt{48 + \sqrt{N}} + N }
\]
Figure \ref{fig:rho_zero} shows a plot of $\rho_0$ as a function of various model vector length $N$.

Since the entire process above is a heuristic based on simplifying assumptions and unidirectional implications, imposing the bound $\rho_0$ on the mean squared elements is too tight, and the actual bound we use should be larger. Consequently, we used $\rho_0$ is used as a starting point for a simple exhaustive search: we trained ten models with $\rho \in \{ k \rho_0 : k = 1, \dots, 10 \}$, and selected the smallest $k$ for which models trained with a larger $k$ show no visible log-loss improvement.

\section{Evaluation}\label{sec:evaluation}
Our aim was reducing the number of instances discarded by the hyper-parameter tuning mechanism while improving the model's accuracy. We therefore evaluate the performance of our method along these two criteria. 

\begin{figure*}[htpb]
    \centering
    \includegraphics[width=.9\textwidth]{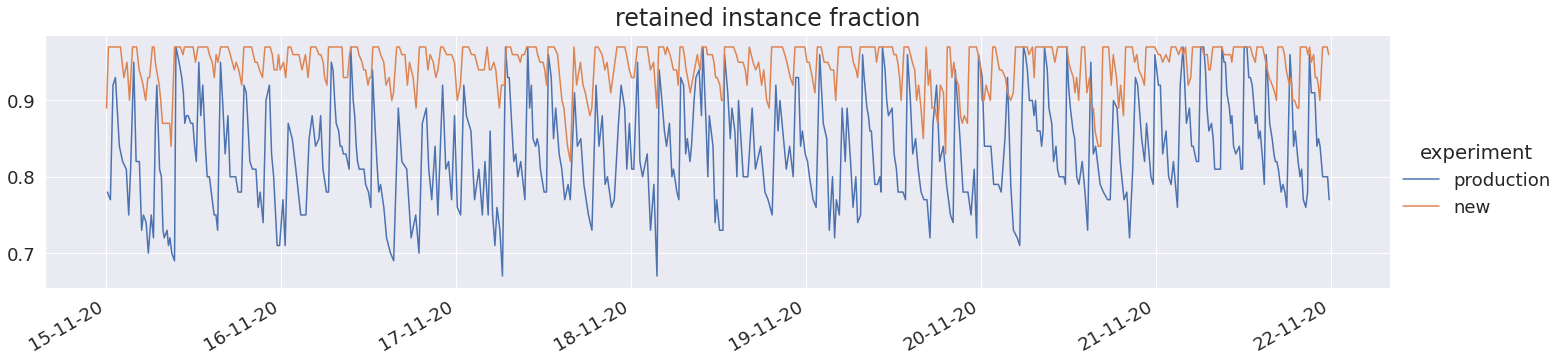}
    \caption{The fraction of retained instances in each training cycle over the course of a week. The $y$ axis is the fraction, between 0 and 1, of the instances that were retained in each training cycle. The orange line is our improved algorithm, whereas the blue line is the existing production algorithm.}
    \label{fig:divergence_ratio}
\end{figure*}

\begin{figure*}[htpb]
    \centering
    \includegraphics[height=.16\textheight]{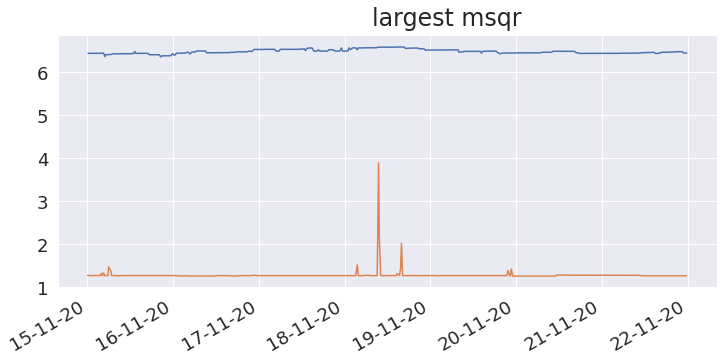}
    \hspace{.05\textwidth}
    \includegraphics[height=.16\textheight]{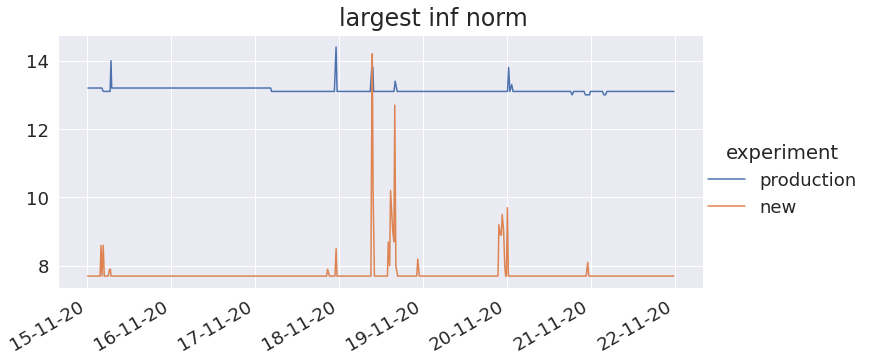}
    \caption{The largest mean squared element and infinity norm of any vector encountered in the model, as a function of time, measured for both the production and the new model. A substantial improvement is observed for both measures.}
    \label{fig:norm_control}
\end{figure*}

We measured, over the course of a week, the fraction, between 0 and 1, of retained instances in each training cycle. The results are plotted in Figure \ref{fig:divergence_ratio}, where the orange line depicts our improved algorithm, whereas the blue line depicts the algorithm deployed in production at the time. It is apparent that the production algorithm often discards over 20 percent of its training instances, whereas our improved algorithm rarely discards more than a few percent.
We also measured the capability of our algorithm to actually keep vector norms under control. To that end, we plotted the maximum infinity norm and mean squared element among the model's latent vectors after each training cycle for both the new and the production model. The results are in Figure \ref{fig:norm_control}, where we see a vast improvement over the production model. Moreover, we see that indeed controlling the mean squared element also keeps the infinity norm in check.

\begin{figure}[htpb]
\includegraphics[width=\columnwidth]{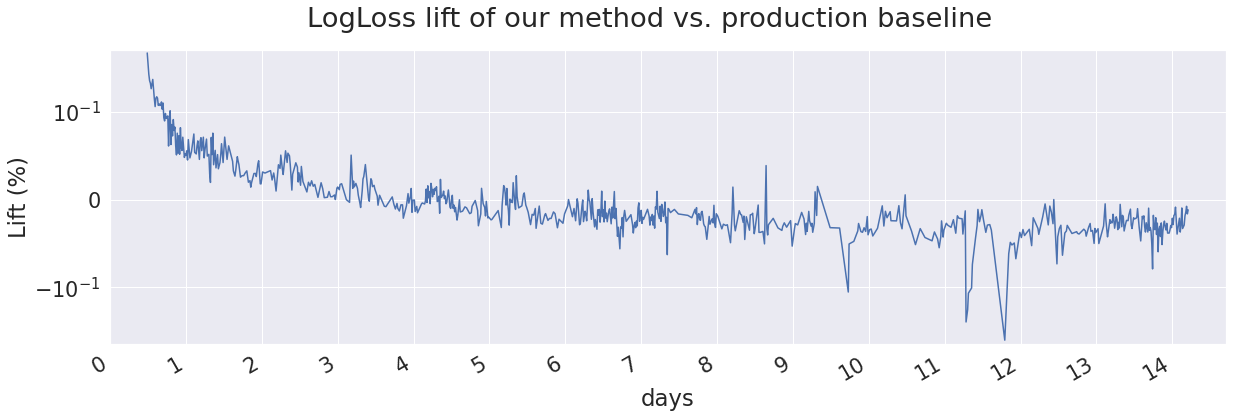}
\caption{LogLoss lift between our algorithm and the production algorithm over the course of two weeks. The model our algorithm trained was initialized from the production model. We can see that except for a short "acclimation" period during the first day, the new algorithm consistently improves the LogLoss metric.}
\label{fig:logloss_divmit_prod}
\end{figure}

To back the claim that, except for reducing the number of diverging instances, our method improves model accuracy, we measured the LogLoss lift between a model training with our algorithm and our production algorithm. In Figure \ref{fig:logloss_divmit_prod} we can see that except for a short "acclimation" period during the first day, the new algorithm consistently improves the LogLoss metric.

We also conducted an online A/B test comparing our new algorithm to the one deployed in production, in terms of revenue impact measured by CPM (the cost of 1000 ad views), and fitting to user preferences terms of CTR. The results are summarized in Table \ref{tab:online_comparison}. The table shows  the relative daily \emph{lifts} observed on each metric by the new model relatively to the production model, e.g. CPM lift is defined as $100 \times \left( \mathrm{\frac{CPM_{new}}{ CPM_{baseline}}} - 1 \right)$. We see an over-all $0.67\%$ CPM lift and $0.93\%$ CTR lift, which are quite significant for an online advertising product. 

\begin{figure}[htpb]
\centering
\includegraphics[width=.48\columnwidth]{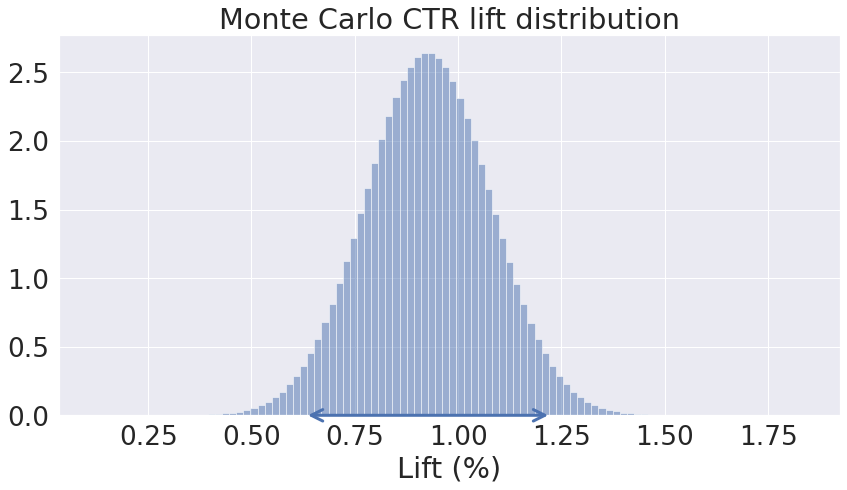}
\includegraphics[width=.48\columnwidth]{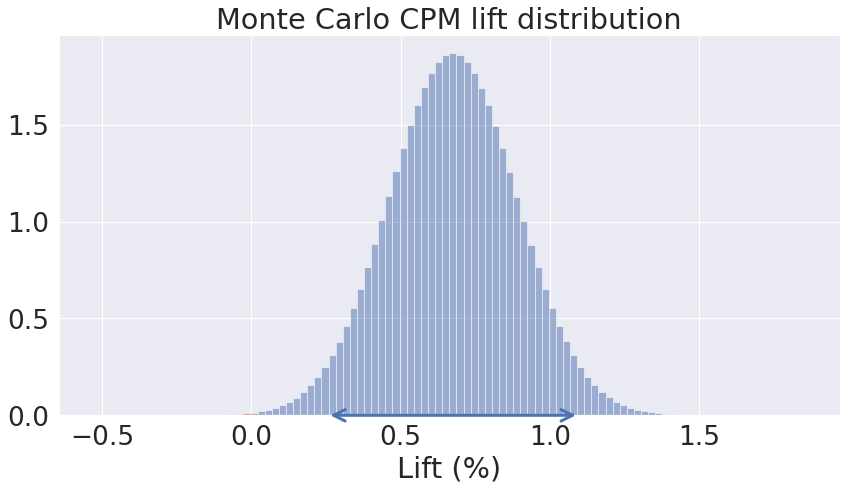}
\caption{Monte-Carlo simulation of CPM and CTR lifts, generated by sampling the Bayesian posteriors. 95\% credible intervals are drawn on the $x$ axis using blue arrows. The CTR credible interval is $[0.63, 1.02]$, whereas the CPM credible interval is $[0.26, 1.09]$. }
\label{fig:uncertainty}
\end{figure}

We measure the reliability of the overall lifts using a Bayesian A/B testing methodology similar to \cite{stucchio2015bayesian}, since it allows to model the revenue generation process. For each experiment, costs per click are modeled using an exponential distribution with parameter $\lambda$, while CTR is modeled using a Bernoulli distribution with parameter $p$. Consequently, CPM is modeled by $\mathrm{CPM} = 1000 \times \tfrac{p}{\lambda}$. The lift comparing CTR and CPM of both experiments are defined accordingly. By constructing a posterior distribution of $p$ and $\lambda$ for both experiments, we use Monte-Carlo simulation to compute two the 95\% credible interval $[l, u]$, such that $l, u$ corresponds the $2.5\%$ and $97.5\%$ percentiles, respectively. Looking at Figure \ref{fig:uncertainty} and the corresponding credible intervals, it is apparent that the overall lifts are strictly positive.

\begin{table}[htpb]
    \centering
    \begin{tabular}{|c|c|c|}
        \hline
        Day & CPM Lift (\%) & CTR Lift (\%) \\
        \hline
        1 & 0.94 & 0.64 \\
        \hline
        2 & 0.13 & -0.2  \\
        \hline
        3 & 0.56 & 0.14 \\
        \hline
        4 & 1.35 & 0.91 \\
        \hline
        5 & 0.24 & 0.86 \\
        \hline
        6 & -0.67 & 0.58 \\
        \hline
        7 & 1.54 & 2.09 \\
        \hline
        8 & -0.27 & 1.68 \\
        \hline
        9 & 2.16 & 1.6 \\
        \hline
    \end{tabular}
    \caption{An online comparison between our improved algorithm and its baseline, the production algorithm.}
    \label{tab:online_comparison}
\end{table}

\begin{figure}[htpb]
\centering
\includegraphics[width=\columnwidth]{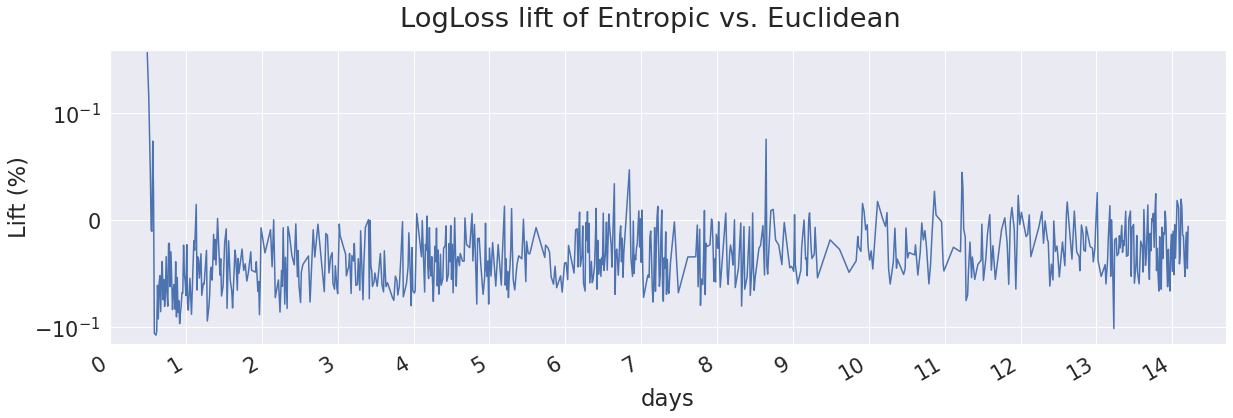} \\
\includegraphics[width=\columnwidth]{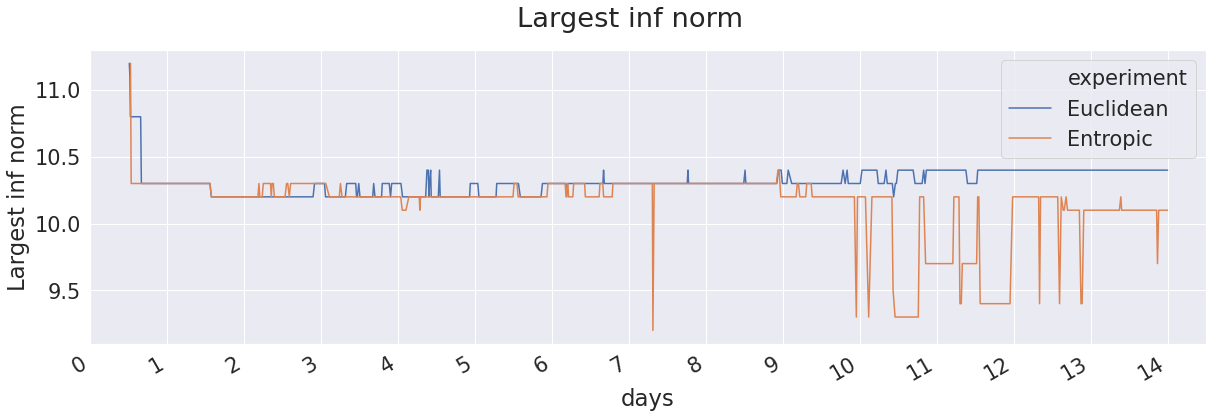} \\
\includegraphics[width=\columnwidth]{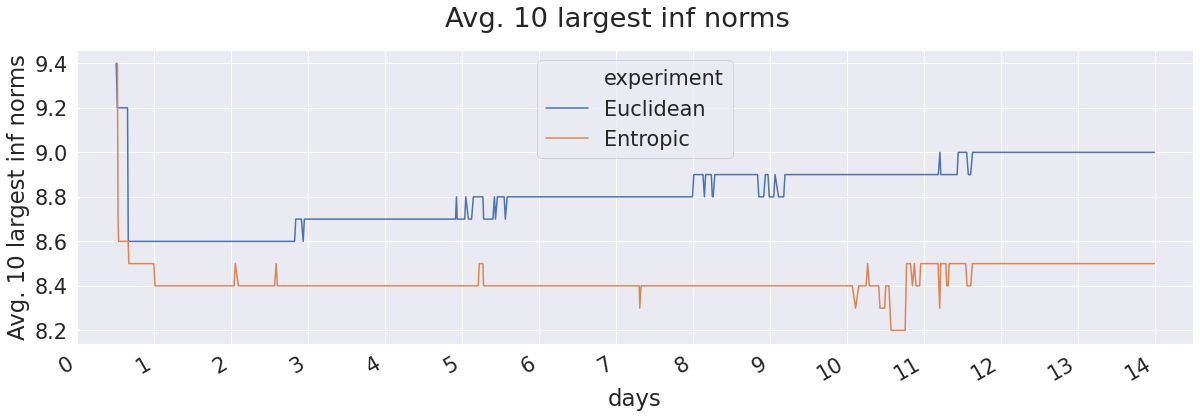} \\
\caption{Comparison of entropic ascent to naive Eucldean gradient ascent. Top - LogLoss lift. Middle - largest infinity norm among the vectors in the model. Bottom - average of 10 largest infinity norms of vectors in the model. It is apparent that the first few training rounds are an "acclimation" period where the LogLoss difference stabilizes and the inifnity norms drop. Afterwards, we see that the entropic algorithm consistently improves LogLoss while maintaining lower vector infinity norms.}
\label{fig:euclidean}
\end{figure}

Finally, we show that the multiplicative updates in equation \eqref{eq:entropic_ascent} are indeed better than the naive projected gradient ascent algorithm, which stems from using a Euclidean distance measure in the proximal formulation. We plotted in Figure \ref{fig:euclidean} the results of comparing the entropic algorithm to the euclidean algorithm with the best-performing step-size, in terms of LogLoss, we could achieve. We can see that the entropic algorithm is both better at controlling vector norms, \emph{and} at achieving a better LogLoss.

\bibliographystyle{plain}
\bibliography{references}

\end{document}